\documentclass[lettersize,journal]{IEEEtran}
\usepackage{amsmath,amsfonts}
\usepackage{array}
\usepackage[caption=false,font=normalsize,labelfont=sf,textfont=sf]{subfig}
\usepackage{textcomp}
\usepackage{stfloats}
\usepackage{url}
\usepackage{verbatim}
\usepackage{graphicx}
\usepackage{cite}
\usepackage[ruled,linesnumbered]{algorithm2e}
\usepackage{siunitx}
\SetKwInput{KwInit}{Initialize}
\usepackage{booktabs} 
\SetKwInput{KwInst}{Instantiate}
\usepackage[hidelinks]{hyperref}
\usepackage{cancel}

\hypersetup{
    colorlinks=true,
    urlcolor=blue,
    citecolor=black,
    linkcolor=black
}
\begin{document}

\title{EditFollower: Tunable Car Following Models for Customizable Adaptive Cruise Control Systems}

\author{Xianda Chen, Xu Han, Meixin Zhu, Xiaowen Chu, PakHin Tiu, Xinhu Zheng, Yinhai Wang,~\IEEEmembership{Fellow,~IEEE}
\thanks{This study is supported by the National Natural Science Foundation of China under Grant 52302379, Guangzhou Basic and Applied Basic Research Projects under Grants 2023A03J0106 and 2024A04J4290, Guangdong Province General Universities Youth Innovative Talents Project under Grant 2023KQNCX100, and Guangzhou Municipal Science and Technology Project 2023A03J0011.

(Xianda Chen and Xu Han contributed equally to this work.) (Corresponding authors: Meixin Zhu, Xiaowen Chu)}
\thanks{Xianda Chen, Meixin Zhu, PakHin Tiu, Xinhu Zheng are with the Intelligent Transportation Thrust, Systems Hub, The Hong Kong University of Science and Technology (Guangzhou), Guangzhou, China; Meixin Zhu is also with Guangdong Provincial Key Lab of Integrated Communication, Sensing and Computation for Ubiquitous Internet of Things (email: xchen595@connect.hkust-gz.edu.cn, meixin@ust.hk, phtiu454@connect.hkust-gz.edu.cn, xinhuzheng@hkust-gz.edu.cn).}%

\thanks{Xu Han and Xiaowen Chu are with the Data Science and Analytics Thrust, Information Hub, The Hong Kong University of Science and Technology (Guangzhou), Guangzhou, China (email: xhanab@connect.ust.hk, xwchu@ust.hk).}%
\thanks{Yinhai Wang is with the Department of Civil
and Environmental Engineering, University of Washington, Seattle, WA 98195 USA (e-mail: yinhai@uw.edu).}
}

\maketitle

\begin{abstract}
In the realm of driving technologies, fully autonomous vehicles have not been widely adopted yet, making advanced driver assistance systems (ADAS) crucial for enhancing driving experiences. Adaptive Cruise Control (ACC) emerges as a pivotal component of ADAS. However, current ACC systems often employ fixed settings, failing to intuitively capture drivers' social preferences and leading to potential function disengagement. To overcome these limitations, we propose the Editable Behavior Generation (EBG) model, a data-driven car-following model that allows for adjusting driving discourtesy levels. The framework integrates diverse courtesy calculation methods into long short-term memory (LSTM) and Transformer architectures, offering a comprehensive approach to capture nuanced driving dynamics. By integrating various discourtesy values during the training process, our model generates realistic agent trajectories with different levels of courtesy in car-following behavior. Experimental results on the HighD and Waymo datasets showcase a reduction in Mean Squared Error (MSE) of spacing and MSE of speed compared to baselines, establishing style controllability. To the best of our knowledge, this work represents the first data-driven car-following model capable of dynamically adjusting discourtesy levels. Our model provides valuable insights for the development of ACC systems that take into account drivers' social preferences.
\end{abstract}

\begin{IEEEkeywords}
Autonomous Vehicles, Advanced Driver Assistance Systems, Adaptive Cruise Control, Car-Following.
\end{IEEEkeywords}

\section{Introduction}
\IEEEPARstart{I}n the current landscape of driving technologies, fully autonomous vehicles are yet to become ubiquitous. Advanced driver assistance systems (ADAS), consequently, still play an important role in driving experience enhancement. Among these, Adaptive Cruise Control (ACC) stands out as a cornerstone, serving as the most crucial and prevalent component of ADAS. 

Accurately modeling car-following behavior is paramount for micro-level traffic simulation and plays a pivotal role in ACC systems. Researchers have shown a keen interest in developing car-following models over the years to simulate and comprehend the dynamic nature of car-following behavior \cite{treiber2000congested, wiedemann1974simulation, wang2016drivers, wang2022effect, chen2023follownet,chen2023bayesian}. 

However, existing ACC systems often utilize fixed settings like predetermined distances or time headways \cite{makridis2019response}. These standardized approaches fall short of satisfying the heterogeneous social preferences of drivers. Recognizing that drivers have distinct driving styles \cite{wang2018capturing, wang2018driving}, the inability to cater to these preferences may lead to disengagement with ACC systems \cite{parasuraman1997humans, larsson2012driver}, limiting their effectiveness in optimizing driving experiences. The development of an ACC system that allows for adjustable driving styles could significantly enhance acceptance and user experience.

Moreover, despite the promise of deep learning methods for replicating real-world driving behavior \cite{li2022metadrive, suo2021trafficsim, bergamini2021simnet}, these approaches remain confined to mimicking the training data's distribution. This hinders their ability to control or deviate from observed patterns, impeding the exploration of diverse and potentially safer driving strategies.

\begin{figure}
\centering
\includegraphics[width=3.4in]{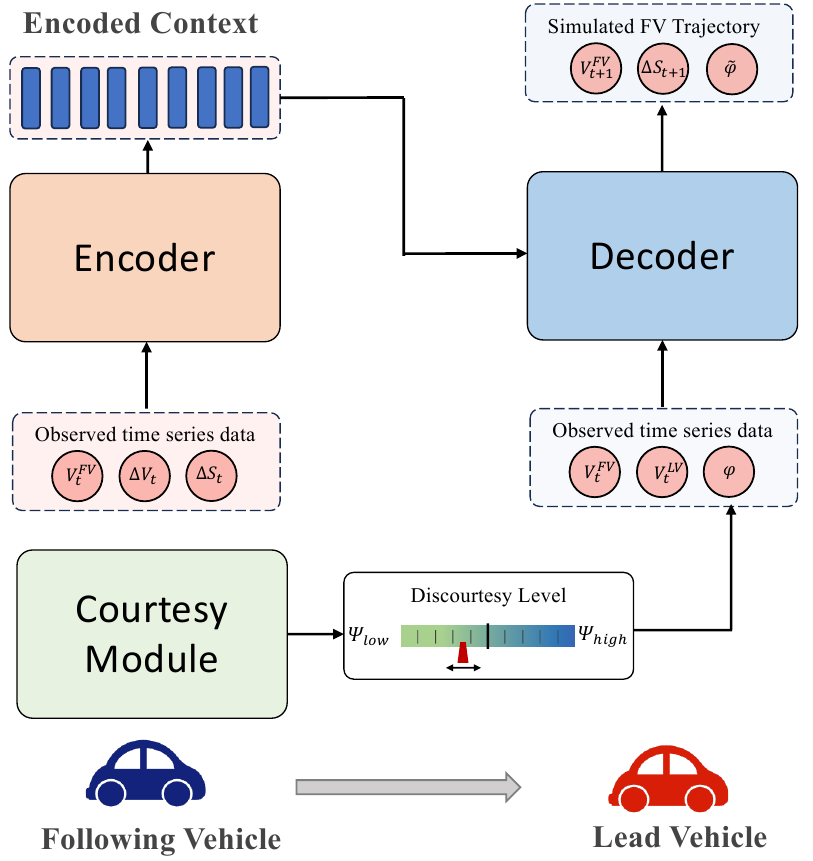}
\caption{Overview: simulating tunable car following
behaviors. The model utilizes real-world car-following data to synthesize a simulated following vehicle's future trajectory based on an input discourtesy value. }

\label{fig:over_model}
\end{figure}

To address these limitations and better simulate car-following behaviors in various interactive traffic scenarios, we propose a comprehensive approach. We first integrate diverse discourtesy calculation methods into state-of-the-art data-driven car-following models, including long short-term memory (LSTM) and Transformer architectures. Inspired by previous work \cite{sun2018courteous, schwarting2019social, chang2023editing, zhong2023guided}, we then introduce the Editable Behavior Generation (EBG) model. This model represents the first data-driven car-following approach that allows for the adjustment of driving discourtesy level,  as shown in Fig. \ref{fig:over_model}.

Specifically, we utilize real-world car-following data with discourtesy labels to train the EBG model. This model synthesizes a simulated following vehicle's future trajectory based on an input discourtesy value, allowing for the generation of realistic car-following behaviors. We present diverse frameworks to label the discourtesy value during training and design a courtesy loss that matches the discourtesy values of the generated trajectories with the input values. The effectiveness of our model is validated using the HighD and Waymo datasets. Results demonstrate that our proposed method is effective in generating desired driving behaviors. Additionally, it captures different driving styles, leading to a reduction in the Mean Squared Error (MSE) of spacing and the MSE of speed. Furthermore, our findings offer valuable insights for the development of Adaptive Cruise Control functions. By facilitating the adjustment of discourtesy levels, our model aligns more closely with drivers' social preferences, thereby enhancing the applicability and effectiveness of ACC technologies in real-world scenarios.

Our main contributions are as follows:
\begin{itemize}
\item We propose the first data-driven car-following model that can simulate driving behavior according to discourtesy levels.
\item We demonstrate the control ability of our model in generating realistic car-following behaviors on the naturalistic datasets.
\item Our model can be seamlessly integrated into any data-driven car-following model, enhancing its versatility and applicability while providing valuable insights for advancing Adaptive Cruise Control functions.
\end{itemize}

\section{Related Work} \label{section:related}
\subsection{Rule-based car-following models}
Since the first car-following model, the Pipes model, was proposed in 1953 \cite{pipes1953operational}, researchers have devoted over seven decades to studying car-following models. Five main categories of traditional car-following models have been identified: stimulus-response models, safe distance models, psycho-physiological models, optimal velocity models, and desired goal models \cite{saifuzzaman2014incorporating, olstam2004comparison, chen2023follownet}.

Stimulus-response models rely solely on relative information (distance or speed) to predict FV acceleration. Built upon the principle of direct response to speed difference, the Gazis-Herman-Rothery (GHR) model \cite{chandler1958traffic} assumes a vehicle's acceleration as directly proportional to the speed difference between itself and the lead vehicle, neatly capturing its basic acceleration/deceleration dynamics.

Different from stimulus-response models, Kometani $et$ $al.$ \cite{kometani1959dynamic} introduced the safety distance model, acknowledging the inherent unpredictability of lead vehicles. This "conflict avoidance" model prioritizes maintaining a minimum safe distance from the lead vehicle, setting a new standard for car-following behavior.

The psycho-physiological model, built on Michaels' work \cite{michaels1963perceptual}, supposes that drivers react primarily to changes in relative LV-FV motion. VISSIM\ensuremath{^{{\textsuperscript{\textregistered}}}}, a microscopic traffic simulator, leverages this concept through the Wiedemann model \cite{wiedemann1974simulation}.

While the Optimal Velocity Model (OVM) of Bando $et$ $al.$ \cite{bando1995dynamical} effectively explains diverse traffic flow phenomena, its potential for unrealistic acceleration/deceleration prompted the development of the Generalized Force Model (GFM) by Helbing $et$ $al.$ \cite{helbing1998generalized}. This model augments the OVM by incorporating the influence of negative speed differences, leading to more realistic driving behaviors. 

Driver-centric desired goal models, assuming specific goals like desired speed and headway, have become popularized by models like the Intelligent Driver Model (IDM) \cite{treiber2000congested}. This stands out as the most widely used traditional car-following model which offers a simpler and more effective approach to modeling driver behavior. 

However, it is important to note that these rule-based models generally focus on factors such as relative distance, speed difference, and safety distance, they do not explicitly incorporate the concept of courtesy into their frameworks. Courtesy, which refers to the willingness of drivers to accommodate and show consideration for others, plays a crucial role in shaping driver behavior and overall traffic flow.

\subsection{Data-driven car-following models}
In the field of car-following models, there are two main categories: traditional models and data-driven models. Data-driven car-following models transcend traditional rule-based approaches by embracing the power of Artificial Intelligence. From nonparametric regression's flexibility to the intricate learning capabilities of neural networks and reinforcement learning, these models are trained from rich datasets, unraveling the intricate tapestry of factors influencing driver behavior.

For example, in 2015, a simple k-nearest neighbor-based nonparametric car-following model was introduced to predict driving behavior \cite{he2015simple}. Another nonparametric data-driven model is the Loess model, which relies on locally weighted regression \cite{papathanasopoulou2015towards}.

A 4-layer neural network structure with two hidden layers, as adopted by Hongfei $et$ $al.$ \cite{hongfei2003develop}, demonstrates the effectiveness of neural network-based approaches in capturing the complex relationships between gap distance, relative velocity, desired velocity, and FV speed. Reactive agent-based models employing ANNs, such as backpropagation and fuzzy ARTMAP \cite{panwai2007neural}, have emerged as promising approaches in car-following research due to their ability to learn complex traffic dynamics and adapt to diverse driving scenarios. Additionally, research has demonstrated the effectiveness of recurrent neural networks, including Vanilla RNN, Long Short-Term Memory (LSTM), and Gated Recurrent Unit (GRU), in modeling car-following behavior \cite{zhou2017recurrent, ma2020sequence, wang2017capturing}. Also, Mo $et$ $al.$ \cite{mo2021physics} introduced the Physical Information Deep Learning Car-Following (PIDL-CF) model. This approach integrates traditional car-following models with neural networks to predict acceleration under different traffic conditions. 

In the realm of reinforcement learning (RL), several frameworks based on deep deterministic policy gradient (DDPG) have been developed to model car-following behavior, aiming to replicate human-like behavior while optimizing driving performance \cite{zhu2018human, zhu2020safe, han2023ensemblefollower}. Furthermore, personalized car-following models have emerged using inverse reinforcement learning (IRL) \cite{gao2018car} or hybrid approaches combining RL and supervised learning \cite{zhao2022personalized}, allowing for individual driver behavior modeling.


Recently, attention has been drawn to the potential of using Transformer-based models for long-sequence car-following trajectory prediction \cite{zhu2022transfollower,fang2023dynamic}. These approaches excel at capturing complex temporal relationships within driving data. Despite their ability to learn reactive driving from vast datasets, deep learning methods remain tethered to the distribution of training data. This limits their capacity to modify agent courtesy beyond the spectrum observed in the training data. 

\subsection{Controllable driving behavior}

Extensive research has been conducted on the development of Adaptive Cruise Control systems that mimic human driving behavior, followed by a substantial body of work in the field of controllable driving behavior. Zhu $et$ $al.$  \cite{zhu2019typical} introduced a data-driven approach to create a Personalized Adaptive Cruise Control system capable (PACC) of adapting to different driving styles. This method involves gathering real-world driving data, categorizing drivers based on their unique characteristics, and then constructing a personalized PACC architecture. Similarly, Lu $et$ $al.$  \cite{lu2019personalized} proposed a personalized behavior learning system (PBLS) using neural reinforcement learning. By mimicking the nuanced adjustments of experienced drivers, the PBLS transcends the limitations of conventional motion planners, unlocking a new era of human-centric longitudinal control in diverse driving scenarios. In addition, Wei $et$ $al.$ \cite{wei2023mpc} used simulator data to fine-tune a car-following model that imitates human driving behavior and developed a Model Predictive Control controller. This controller allows autonomous vehicles to smoothly follow natural trajectories in various situations. However, accurately replicating the entire spectrum of human driving behaviors presents a significant challenge with these methods due to their reliance on large-scale real-world driving data.

Achieving a thorough emulation of human driving behavior entails capturing not only the diverse range of driving styles and scenarios but also detecting the nuanced decision-making processes and situational awareness that human drivers possess. Therefore, it is important to enhance the adaptability and responsiveness of these systems across a wider range of driving conditions and individual driving preferences. Chang $et$ $al.$ \cite{chang2023editing} proposed an innovative strategy called Socially-Controllable Behavior Generation (SCBG). This strategy introduces a new parameter, namely the courtesy level $\psi$, to specifically address how courteously Vehicle B behaves towards Vehicle A in a simulated interactive traffic scenario. The SCBG model proved its mettle on the Waymo Open Motion Dataset. Not only did it generate realistic driving behaviors with adjustable courtesy levels, but it also demonstrated the remarkable ability to adapt courteous actions to the specific demands of each scenario. Therefore, the concept of courtesy level is applied to generate various car-following behaviors in our research, modeling the diverse following behaviors arising from different driver profiles, from aggressive tailgating to cautious distance-keeping, specifically for simulation agents under different levels of aggressiveness.

\section{Editable Behavior Generation  Model}

The Editable Behavior Generation (EBG) Model comprises several key components aimed at simulating the future behavior of a following vehicle based on input courtesy features and historical driving context. The overall methodology can be divided into the following parts:

\subsection{Problem formulation}\label{section:definition}
We define the problem by outlining the input features and output variables of the model. The five dimensions of the input feature are the speed of LV ($v_t^{LV}$), the speed of FV ($v_t^{FV}$), the relative spacing ($\Delta d_t$), the relative speed ($\Delta v_t$), and FV's discourtesy level (denoted as $\psi$). The model's output is the speed of FV in the next time step. The composition of the input features and output variables is listed below. 

\begin{equation}
    Input = x_t = [\Delta d_t, v_t^{LV},v_t^{FV}, \Delta v_t, \psi]  \quad (\psi \in \Psi) 
\end{equation}

\begin{equation}
     Output = \hat{y_t} = [v_{t+1}]
\end{equation}
where $\hat{y_t}$ represents the speed prediction of FV at the next time step.

\subsection{Driving discourtesy level}
To capture diverse driving behaviors, three discourtesy level calculation methods are explored: acceleration-based courtesy, jerk-based courtesy, and speed-based courtesy. Each method provides a metric that influences how the model adjusts its predictions based on the driving context.

\subsubsection{Acceleration-based courtesy}
Acceleration measures the rate of change of a vehicle's velocity and serves as a key indicator of how actively a driver is adjusting their speed. The formula for calculating the discourtesy level based on acceleration metrics is given by:
\[
\text{{Discourtesy Level}} = \frac{{\text{{acc\_std}}}}{{\text{{acc\_mean}}}}
\]

\noindent where,
\begin{itemize}
    \item $\text{{acc\_std}}$: standard deviation of the acceleration values,
    \item $\text{{acc\_mean}}$: mean of the acceleration values.
\end{itemize}

The standard deviation and mean of acceleration provide insights into the variability and average rate of speed changes. A lower standard deviation suggests smoother and more consistent acceleration, indicating a more courteous driving style. On the other hand, a higher mean acceleration may indicate more dynamic driving behavior. The ratio of standard deviation to mean acceleration offers a normalized metric for assessing the aggressiveness of driving. A higher discourtesy level implies a greater variation in acceleration, potentially indicating more aggressive driving tendencies. Considering both the mean and standard deviation of acceleration enables the model to understand how drivers adapt their acceleration profiles. This adaptation is crucial for capturing diverse driving styles in different traffic conditions\cite{langari2005intelligent}.

\subsubsection{Jerk-based courtesy}
Jerk measures the change rate of acceleration. Incorporating jerk metrics as the calculation of discourtesy levels, the car-following model can gain insight into how vehicles adjust their acceleration, reflecting the comfort and perceived aggressiveness of driving behavior. Jerk is particularly relevant in scenarios where smooth acceleration changes are desirable, such as in urban traffic or when considering passenger comfort \cite{feng2017can}.
The formula for calculating the discourtesy level based on jerk metrics \cite{murphey2009driver} is given by:
\[
\text{{Discourtesy Level}} = \frac{{\text{{jerk\_std}}}}{{\text{{jerk\_mean}}}}
\]

\noindent where,
\begin{itemize}
    \item $\text{{jerk\_std}}$: standard deviation of the jerk values,
    \item $\text{{jerk\_mean}}$: mean of the jerk values.
\end{itemize}

\subsubsection{Speed-based courtesy}
Speed is a fundamental parameter influencing driving behavior and can provide valuable insights into the discourtesy level of a driver. The relationship between speed and driving style was investigated in \cite{berry2010effects}. The formula for calculating the discourtesy level based on speed metrics is defined as follows:
\[
\text{{Discourtesy Level}} = \frac{{\text{{speed\_std}}}}{{\text{{speed\_mean}}}}
\]

\noindent where,
\begin{itemize}
    \item $\text{{speed\_std}}$: standard deviation of the speed values,
    \item $\text{{speed\_mean}}$: mean of the speed values.
\end{itemize}

These various discourtesy level metrics contribute to a comprehensive understanding of driving behavior, allowing for nuanced adjustments in driving discourtesy levels.

\subsection{Model architecture}
Our study explores the different combinations of three mainstream car-following models with the three aforementioned discourtesy level calculation methods. Below is the overview of the architecture of each car-following model.

\subsubsection{LSTM model}
The LSTM model is designed with an encoder-decoder structure. The encoder utilizes an LSTM layer to process the input sequence, and the decoder employs another LSTM layer to generate the output sequence. The linear layer maps the hidden states to the final output, which is then scaled and transformed to represent the predicted speed for the future time steps. This model is particularly effective in capturing temporal dependencies in the input data.

\subsubsection{LSTM\_IDM model}
\begin{figure*}[htpb]
\centering
\includegraphics[width = \textwidth]{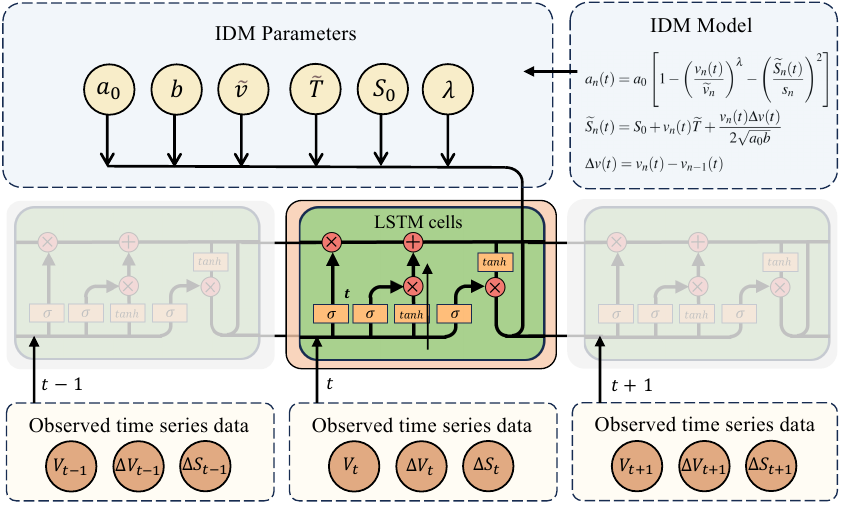}
\caption{LSTM\_IDM Car-following model architecture. Observed time series data, including FV speed, relative speed, and spacing are fed into LSTM cells to capture the temporal dependencies. The LSTM model then uses these extracted features to estimate IDM parameters.}
\label{fig:LSTM_idm}
\end{figure*}

The LSTM\_IDM model extends the LSTM architecture to IDM parameters. In addition to the encoder-decoder structure, this model includes a linear layer that outputs IDM parameters such as desired speed $\widetilde v$, desired time headway $\widetilde{T}$, maximum acceleration $a_0$, comfort acceleration $b$, beta $\lambda$, and jam spacing $S_0$. The IDM parameters are then used to simulate the car-following behavior using IDM equations, yielding predictions for the speed of the FV over time. The overall structure and information flow of the LSTM\_IDM model is shown in Fig. \ref{fig:LSTM_idm}.

\subsubsection{Transformer model}
The Transformer model \cite{vaswani2017attention} features a self-attention mechanism to capture global dependencies in the input sequence. The Transformer's multi-layered architecture, consisting of an encoder and decoder, facilitates parallel processing and scalability for car-following behavior modeling.

\begin{itemize}
    \item $\text{{Encoder}}$: Through multi-head self-attention, the encoder condenses historical driving data into a contextual representation. The input features, including the relative distance (\(\Delta d_t\)), speed (\(v_t\)), relative speed (\(\Delta v_t\)), and discourtesy level(\(\psi\)), are embedded using linear layers. Positional encoding is added to capture temporal information. The resulting encoded input is then fed into the Transformer encoder.
    \item $\text{{Decoder}}$: The decoder generates predictions for the future FV speed based on the speed of the LV and the discourtesy level (\(\psi\)). Similar to the encoder, input features undergo embedding and positional encoding before being fed into the Transformer decoder. The output of the Transformer decoder is passed through a linear layer to produce the final predicted FV speed.
\end{itemize}

\subsection{Loss function}
In the realm of car-following dynamics, the formulation of an effective loss function is paramount for model training and optimization. Given the initial state of the car-following scenario, characterized by the velocity profiles of the LV and the FV, the prediction of the spacing between these vehicles is a critical aspect and is facilitated through the application of system update equations according to Zhu $et$ $al.$ \cite{zhu2020safe}.

The evolution of the system over discrete time intervals $\Delta T$ is encapsulated by the following equations:

\begin{flalign}\label{eq:update}
\begin{split}
&\Delta V(t+1)=V_{LV}(t+1)-V_{FV}(t+1)\\
&S(t+1)=S(t)+\frac{\Delta V(t)+\Delta V(t+1)}{2}*\Delta T
\end{split}
\end{flalign}
where $\Delta T$ represents the simulation time interval, $S$ denotes the spacing between the two vehicles, and $V_{FV}$ and $V_{LV}$ represent the velocities of the following vehicle and lead vehicle, respectively. The loss function for the EBG model is composed of three distinct components, aimed at capturing speed prediction accuracy, spacing accuracy, and courtesy alignment.

\subsubsection{Speed loss (\(L_{\text{speed}}\))}
Measures the similarity between the predicted speed $V_{\text{pred}}$ and the ground truth speed of FV  $V_{\text{label}}$.
    \[ L_{\text{speed}} = \text{criterion}(V_{\text{pred}}, V_{\text{label}}) \]

\subsubsection{Spacing loss (\(L_{\text{spacing}}\))}
Measure the similarity between the predicted spacing (\(S_{\text{pred}}\)) and the actual spacing between the FV and LV (\(S_{\text{label}}\)). The formulation is given by:
\[ L_{\text{spacing}} = \text{criterion}(S_{\text{pred}}, S_{\text{label}}) \]

\subsubsection{Courtesy loss (\(L_{\text{courtesy}}\))}
The courtesy loss is introduced to align the predicted discourtesy levels (\(C_{\text{pred}}\)) with the desired discourtesy levels (\(C_{\text{label}}\)). The formulation is given by:
\[ L_{\text{courtesy}} = \text{criterion}(C_{\text{pred}}, C_{\text{label}}) \]

\subsubsection{Overall loss (\(L_{\text{total}}\))}
 The total loss is a linear combination of the three individual losses, with a tunable weight (\( \alpha \)) assigned to the courtesy loss to emphasize style adherence.
    \[ L_{\text{total}} = L_{\text{speed}} + L_{\text{spacing}} + \alpha \times L_{\text{style}} \]

The optimization process aims to minimize \(L_{\text{total}}\) during training, with \( \alpha \) being a hyperparameter that can be adjusted.

\section{Experiments and Results} 
The experimental evaluation of the proposed EditFollower framework is conducted on the HighD and Waymo datasets.
\subsection{HighD dataset}
\begin{figure}
  \begin{minipage}[t]{0.5\linewidth}
    \centering
    \includegraphics[scale=0.4]{{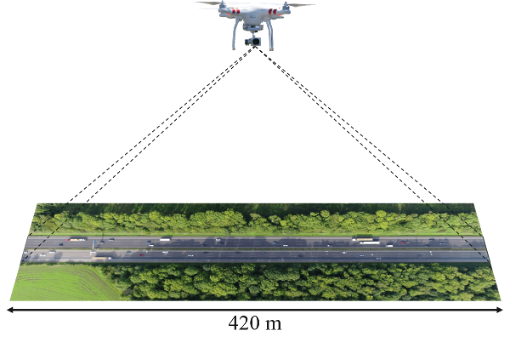}}
\caption{HighD dataset.}
\label{Figure 2}
  \end{minipage}%
  \begin{minipage}[t]{0.5\linewidth}
    \centering
    \includegraphics[scale=0.4]{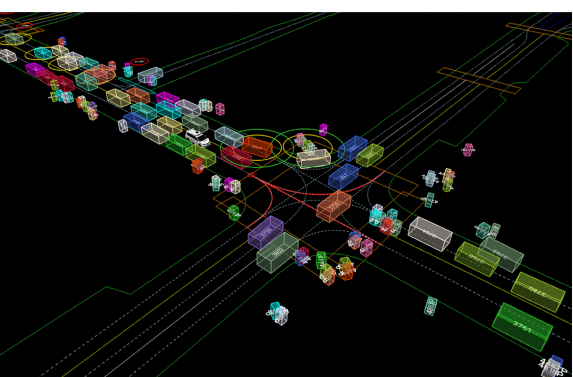}
\caption{Waymo dataset.}
\label{Figure 3}
  \end{minipage}
\end{figure}

HighD dataset \cite{highDdataset} is proposed by the Institute of Automotive Engineering at RWTH Aachen University in Germany, and offers a comprehensive collection of high-precision driving data. This dataset provides in-depth data on vehicle positions and speeds, obtained from bird's-eye view videos of six distinct roads in the Cologne area. Data acquisition was performed using a high-resolution 4K camera mounted on an aerial drone, guaranteeing positional accuracy within 10 cm, as depicted in Fig. \ref{Figure 2}. The dataset employs advanced computer vision techniques to enhance data quality, utilizing Bayesian smoothing to eliminate noise and achieve smooth motion data. It encompasses more than 110,500 vehicles recorded from six distinct locations, providing automatic extraction of vehicle trajectory, size, type, and maneuvers.

\subsection{Waymo dataset}
Waymo, a self-driving car company under Alphabet, has introduced the Waymo Open Dataset \cite{sun2020scalability}, a comprehensive collection of high-resolution sensor data, illustrated in Fig. \ref{Figure 3}. This dataset encompasses lidar and camera data, delivering accurate 3D vehicle poses and annotated object information. With 1950 scenes, each lasting 20 seconds, the Waymo dataset captures diverse driving scenarios, including expressways and urban streets.

\subsection{Car-following event extraction}
Building upon our established car-following benchmark \cite{chen2023follownet}, we extracted car-following events pertinent to this study. The specific criteria are outlined below:

\begin{itemize}
\item The identification number of the lead vehicle should remain constant, indicating that the following vehicle is consistently following the same lead vehicle.

\item The duration of the car-following event should be longer than 15 seconds, guaranteeing that there is enough data to be analyzed for each event.
\end{itemize}

Over 13,980 car-following events (12,540 from HighD and 1,440 from Waymo) were analyzed for this research. To assess model performance under realistic driving conditions, the data was allocated 70/15/15 for training, validation, and testing.

\subsection{Evaluation metrics}

To comprehensively assess the performance of the proposed EditFollower framework, we employ 4 key evaluation metrics. The chosen metrics include Spacing Mean Squared Error (Spacing MSE), Speed Mean Squared Error (Speed MSE), Courtesy Mean Squared Error (Courtesy MSE), and Courtesy Metric Correlation.

\begin{itemize}
    \item \textbf{Spacing MSE}: Measures the mean squared error between predicted and ground truth spacing values. It quantifies the accuracy of spacing predictions.
    
    \item \textbf{Speed MSE}: Quantifies the mean squared error between predicted and actual speeds. This metric assesses the precision of speed predictions.
    
    \item \textbf{Courtesy MSE}: Evaluates the mean squared error between predicted and target discourtesy levels. It gauges the accuracy of discourtesy level predictions.
    
    \item \textbf{Courtesy Metric Correlation}: Examines the correlation between predicted discourtesy levels and labeled discourtesy levels. 
    
\end{itemize}

\begin{table*}[h]
\centering
\caption{Model Performance on HighD Dataset \\("/" means without the integrating courtesy value)}
\label{tab:model_performance}
\begin{tabular}{lcccccc}
\toprule
Model & Spacing MSE & Speed MSE & Courtesy MSE & Courtesy Metric Correlation \\
\midrule
\multicolumn{5}{c}{\textbf{Baseline Models}} \\
\midrule
 LSTM\_IDM & 7.103 & 0.395 & / & / \\
LSTM & 7.314 & 0.436 & / & / \\
Transformer & 7.916 & 0.480 & / & / \\
\midrule
\multicolumn{5}{c}{\textbf{Speed-based Courtesy}} \\
\midrule
   Edit\_LSTM\_IDM & \textbf{6.070} & \textbf{0.283} & \textbf{0.0003} & \textbf{0.972} \\
   Edit\_LSTM & 7.301 & 0.423 & 0.0007 & 0.934 \\
   Edit\_Transformer & 6.360 & 0.388 & 0.0008 & 0.919 \\
\midrule
\multicolumn{5}{c}{\textbf{Acceleration-based Courtesy}} \\
\midrule
   Edit\_LSTM\_IDM & 13.949 & 0.866 & 0.012 & 0.860 \\
   Edit\_LSTM & 8.395 & 0.563 & 0.018 & 0.793 \\
   Edit\_Transformer & 7.731 & 0.676 & 0.009 & 0.972 \\
\midrule
\multicolumn{5}{c}{\textbf{Jerk-based Courtesy}} \\
\midrule
   Edit\_LSTM\_IDM & 12.919 & 0.681 & 0.017 & 0.707 \\
   Edit\_LSTM & 7.556 & 0.436 & 0.031 & 0.423 \\
   Edit\_Transformer & 7.027 & 0.430 & 0.101 & 0.727 \\
\bottomrule
\end{tabular}
\end{table*}

\begin{table*}[t]
\centering
\caption{Model Performance on Waymo Dataset}
\label{tab:waymo_model_performance}
\begin{tabular}{lcccccc}
\toprule
Model & Spacing MSE & Speed MSE & Courtesy MSE & Courtesy Metric Correlation \\
\midrule
\multicolumn{5}{c}{\textbf{Baseline Models}} \\
\midrule
LSTM\_IDM & 11.118 & 0.702 & / & / \\
LSTM & 13.596 & 0.770 & / & / \\
Transformer & 8.450 & 0.726 & / & / \\
\midrule
\multicolumn{5}{c}{\textbf{Speed-based Courtesy}} \\
\midrule
Edit\_LSTM\_IDM & 9.901 & 0.667 & 1.441 & 0.648 \\
Edit\_LSTM & 12.504 & 0.723 & 1.366 & 0.668 \\
Edit\_Transformer & 9.319 & 0.763 & 2.058 & 0.402 \\
\midrule
\multicolumn{5}{c}{\textbf{Acceleration-based Courtesy}} \\
\midrule
Edit\_LSTM\_IDM & \textbf{7.283} & \textbf{0.568} & 0.068 & 0.846 \\
Edit\_LSTM & 13.182 & 0.803 & \textbf{0.040} & 0.900 \\
Edit\_Transformer & 7.573 & 0.979 & 0.051 & \textbf{0.981} \\
\midrule
\multicolumn{5}{c}{\textbf{Jerk-based Courtesy}} \\
\midrule
Edit\_LSTM\_IDM & 10.808 & 0.697 & 0.051 & 0.799 \\
Edit\_LSTM & 12.967 & 0.675 & 0.046 & 0.777 \\
Edit\_Transformer & 8.151 & 0.687 & 0.185 & 0.924 \\
\bottomrule
\end{tabular}
\end{table*}

\section{Model Performance }
The experiment results of various car-following models on both the HighD and Waymo datasets are presented in Tables \ref{tab:model_performance} and \ref{tab:waymo_model_performance}, respectively. These models are evaluated based on the above four metrics. Specifically, models denoted as "LSTM\_IDM," "LSTM," and "Transformer," represent car-following models without the integration of discourtesy levels, serving as baselines for comparison. Therefore, "Courtesy MSE" and "Courtesy Metric Correlation" are denoted as "/" for these models. We highlight the best-performing model results (indicated in \textbf{bold}). 

\subsection{HighD dataset performance}

In speed-based courtesy, The Edit\_LSTM\_IDM model stands out with the lowest Spacing MSE (6.070), Speed MSE (0.283), and Courtesy MSE (0.0003). It achieves a remarkably high Courtesy Metric Correlation of 0.972, indicating strong alignment with desired discourtesy levels.

\subsection{Waymo dataset performance}

In acceleration-based courtesy, The Edit\_LSTM\_IDM model again demonstrates outstanding performance with the lowest Spacing MSE (7.283), Speed MSE (0.568), and Courtesy MSE (0.068). It maintains a high Courtesy Metric Correlation of 0.846, showcasing effective alignment with desired discourtesy levels.

\subsection{Overall insights}

Through analyzing the results presented in the tables, a comparative assessment of different models highlights the effectiveness of our proposed method. The key observations include:
\begin{itemize}
    \item Models incorporating discourtesy levels consistently outperform baseline models across datasets and courtesy calculation methods.
    \item For the HighD dataset, Speed-based courtesy models, particularly Edit\_LSTM\_IDM, demonstrate strong performance in reducing MSE for spacing and speed, as well as effective courtesy alignment. For the Waymo dataset, Acceleration-based courtesy models, Edit\_LSTM\_IDM also exhibit competitive performance.
    \item The Courtesy MSE is minimal, and the Courtesy Metric Correlation is high, demonstrating effective alignment with desired discourtesy levels.
\end{itemize}

These results underscore the effectiveness of the proposed editable behavior generation model in generating realistic car-following behaviors with desired discourtesy levels. The model not only reduces errors in spacing and speed prediction but also ensures a strong correlation between predicted and desired discourtesy levels. The ability to tailor car-following behaviors based on discourtesy levels adds a valuable dimension to the modeling process, contributing to improved accuracy and realism in simulations.

\section{Controllability of Driving Behavior in EBG}
Given the competitive performance of the Edit\_LSTM\_IDM model in terms of Spacing MSE, Speed MSE, and Courtesy MSE on the HighD dataset.
The controllability is examined through a series of experiments conducted on this dataset.

\begin{figure}[h]
\centering
\includegraphics[width=3.4in]{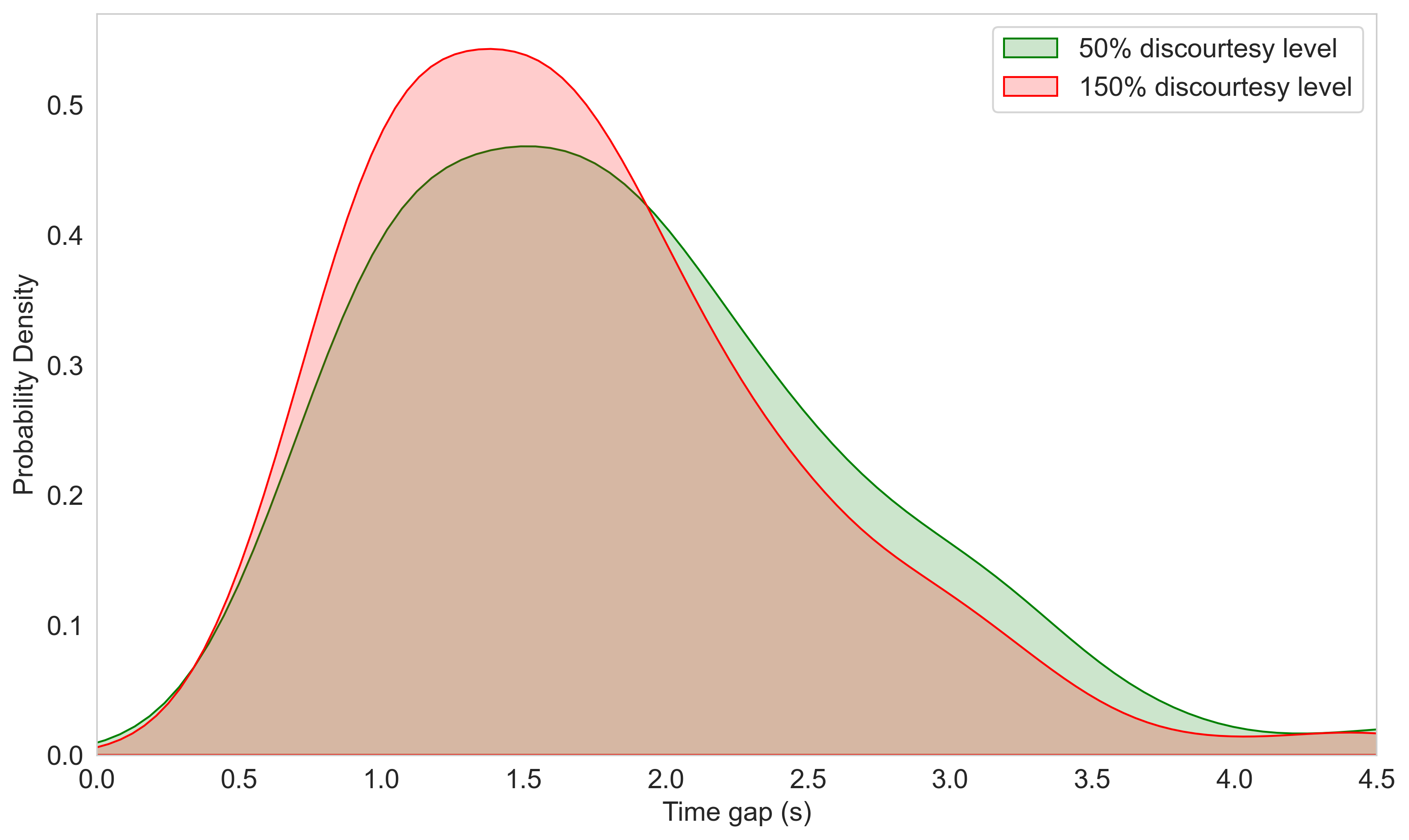}
\caption{Distribution changes when adjusting discourtesy levels. The probability density distribution of the time gap between LV and FV shifts to the left when the discourtesy level changes from 50\% to 150\%, indicating more aggressive driving behaviors.}
\label{fig:1}
\end{figure}

\begin{figure}[h]
\centering
\includegraphics[width=3.4in]{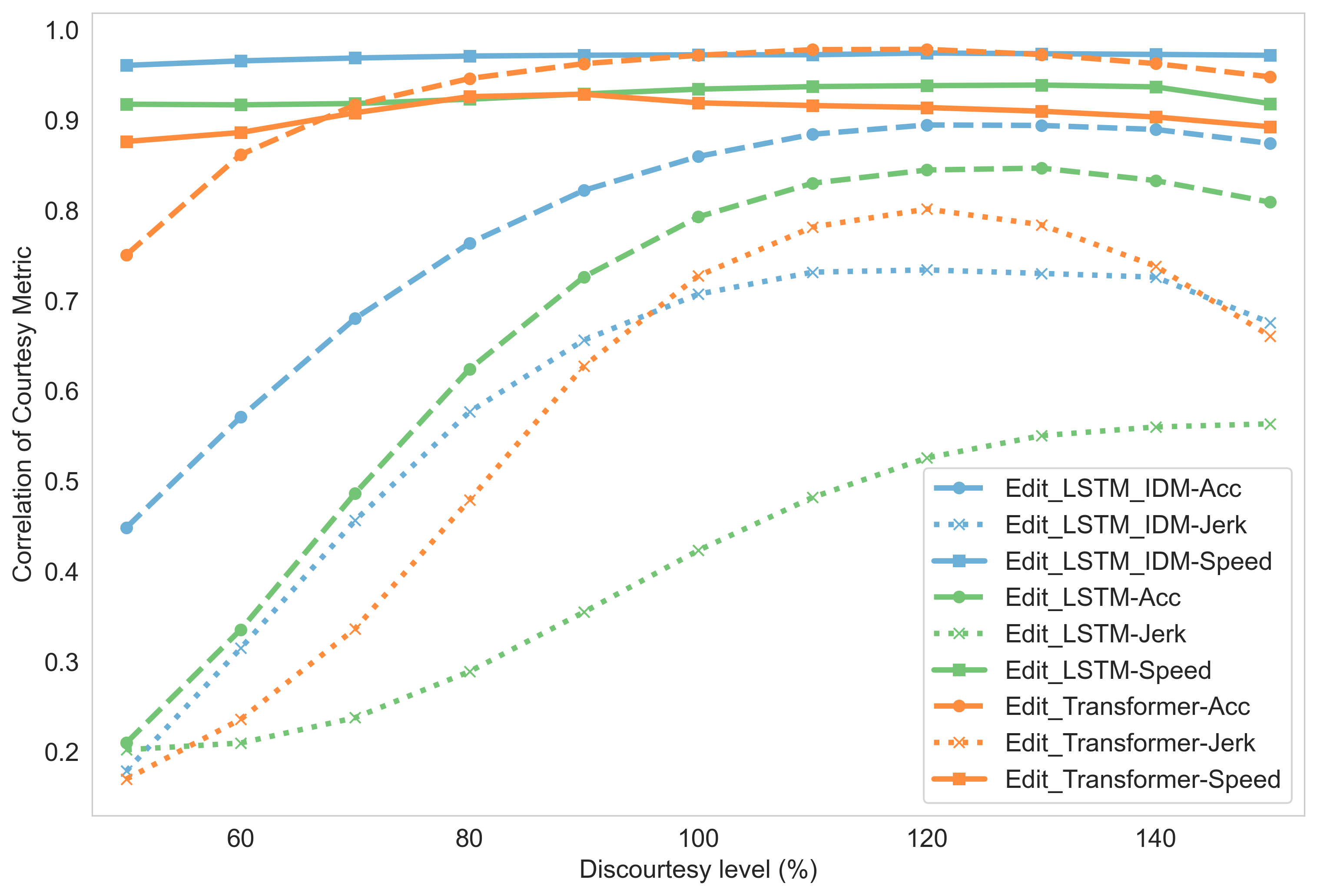}
\caption{Correlation analysis. The correlation between the desired courtesy and the behaved courtesy of different models under multiple discourtesy levels was investigated. A higher correlation value suggests a better match of the desired and behaved courtesy metric.}
\label{fig:2}
\end{figure}    

\subsection{Adjusting discourtesy level}
\begin{enumerate}
    \item Adjusting discourtesy values to 50\% and 150\% of the original, as shown in Fig. \ref{fig:1}, we observe distinct characteristics in the time gap distribution. Specifically, the 150\% discourtesy values exhibit an overall left-skewed distribution, with a higher concentration around 1.4 seconds, indicating more aggressive car-following behavior \cite{xue2019rapid}.
    
    \item Comparative analysis of different input discourtesy values over the entire test set reveals a positive correlation between input discourtesy levels and the corresponding output, as shown in Fig. \ref{fig:2}, this suggests that the model effectively adjusts car-following behavior based on the input courtesy, achieving the desired discourtesy levels.

    \item To provide a more insightful illustration of the influence of discourtesy levels on car-following behavior, Fig. \ref{fig:a} and Fig. \ref{fig:b} showcase variations in speed and spacing derived from real-world data. Additionally, Fig. \ref{fig:4} presents simulated lead vehicle speeds, characterized by a pattern resembling a sine function. Notably, the observed trend indicates that with an increase in discourtesy levels, the following distance diminishes. This reduction implies a transition towards more aggressive car-following behavior, as vehicles tend to close the gap more rapidly.

\begin{figure*}[htbp]
\centering
\begin{minipage}[t]{0.48\textwidth}
\centering
\includegraphics[width = \textwidth]{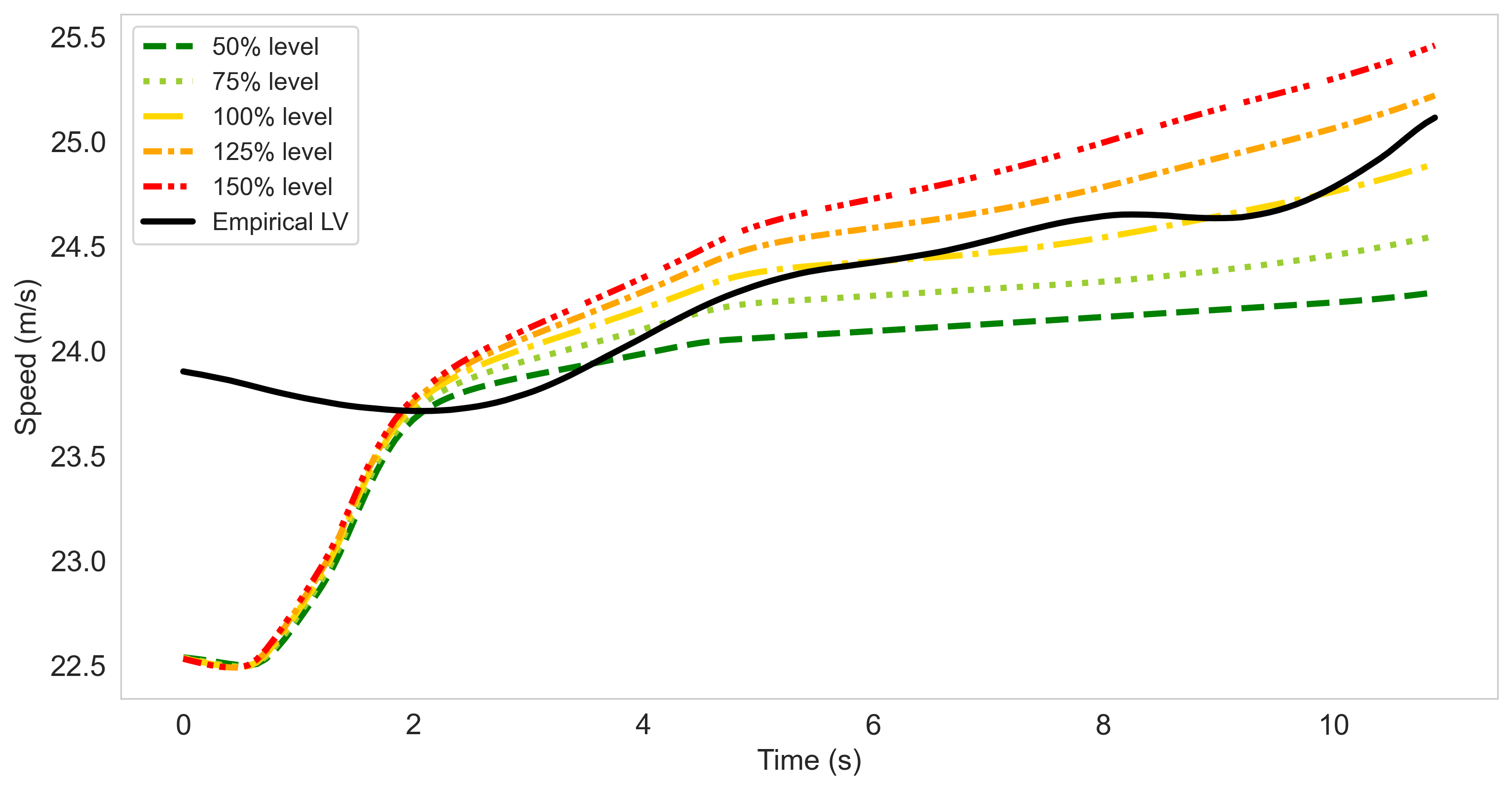}
\end{minipage}
\begin{minipage}[t]{0.48\textwidth}
\centering
\includegraphics[width = \textwidth]{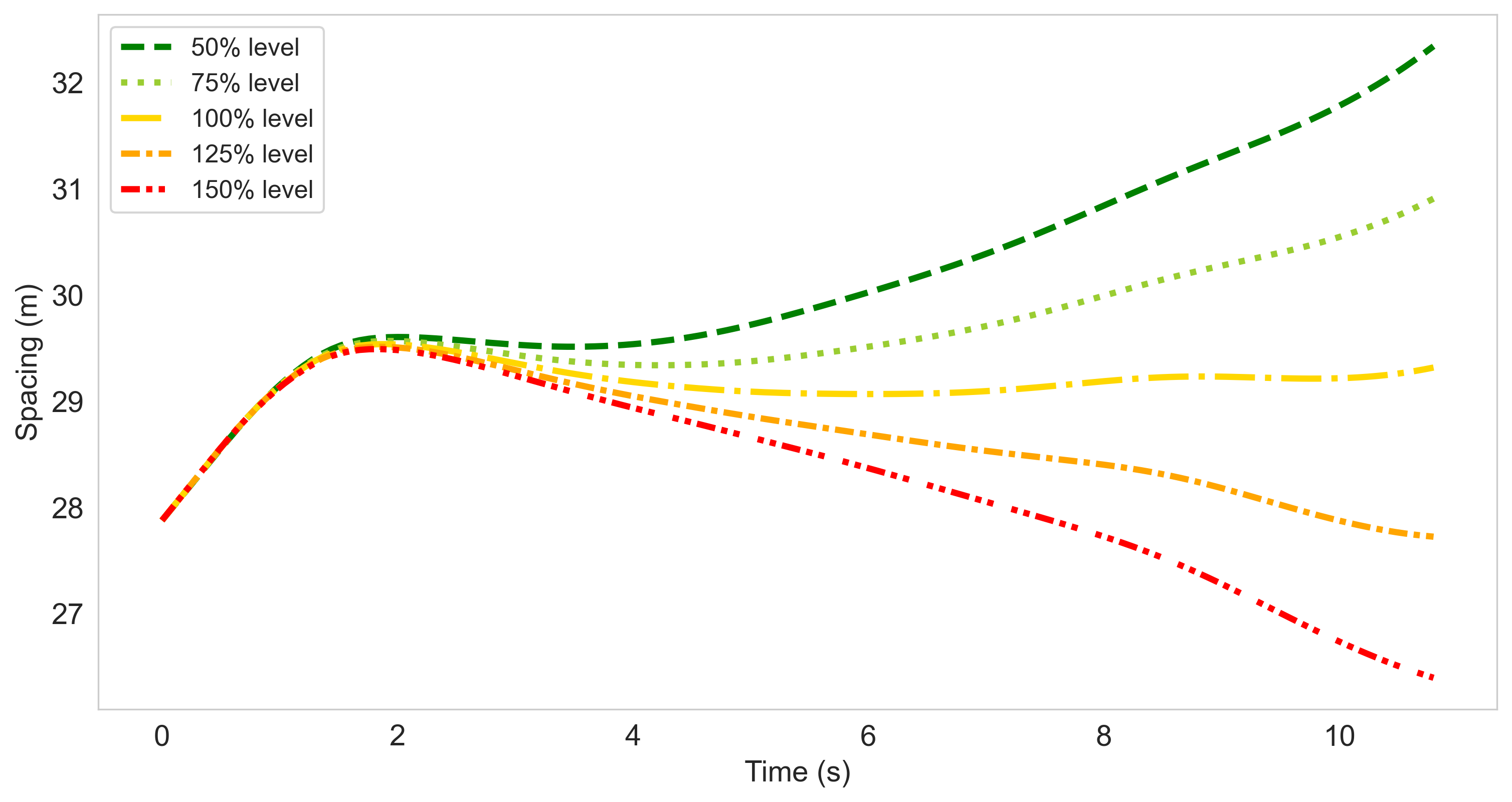}

\end{minipage}
\caption{Variations in car-following behavior with discourtesy levels: case(a).}
\label{fig:a}
\end{figure*}

\begin{figure*}[htbp]
\centering
\begin{minipage}[t]{0.48\textwidth}
\centering
\includegraphics[width = \textwidth]{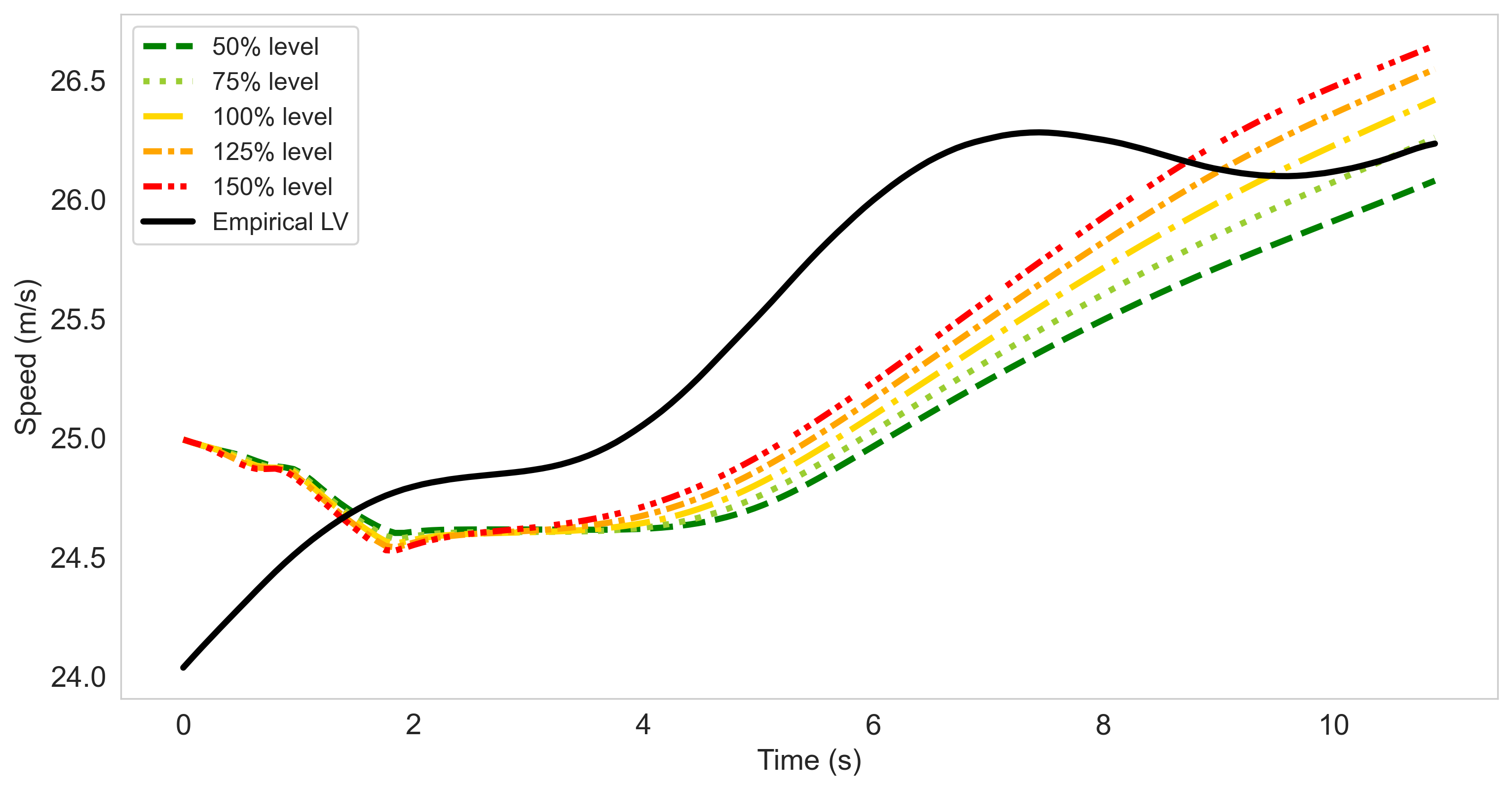}
\end{minipage}
\begin{minipage}[t]{0.48\textwidth}
\centering
\includegraphics[width = \textwidth]{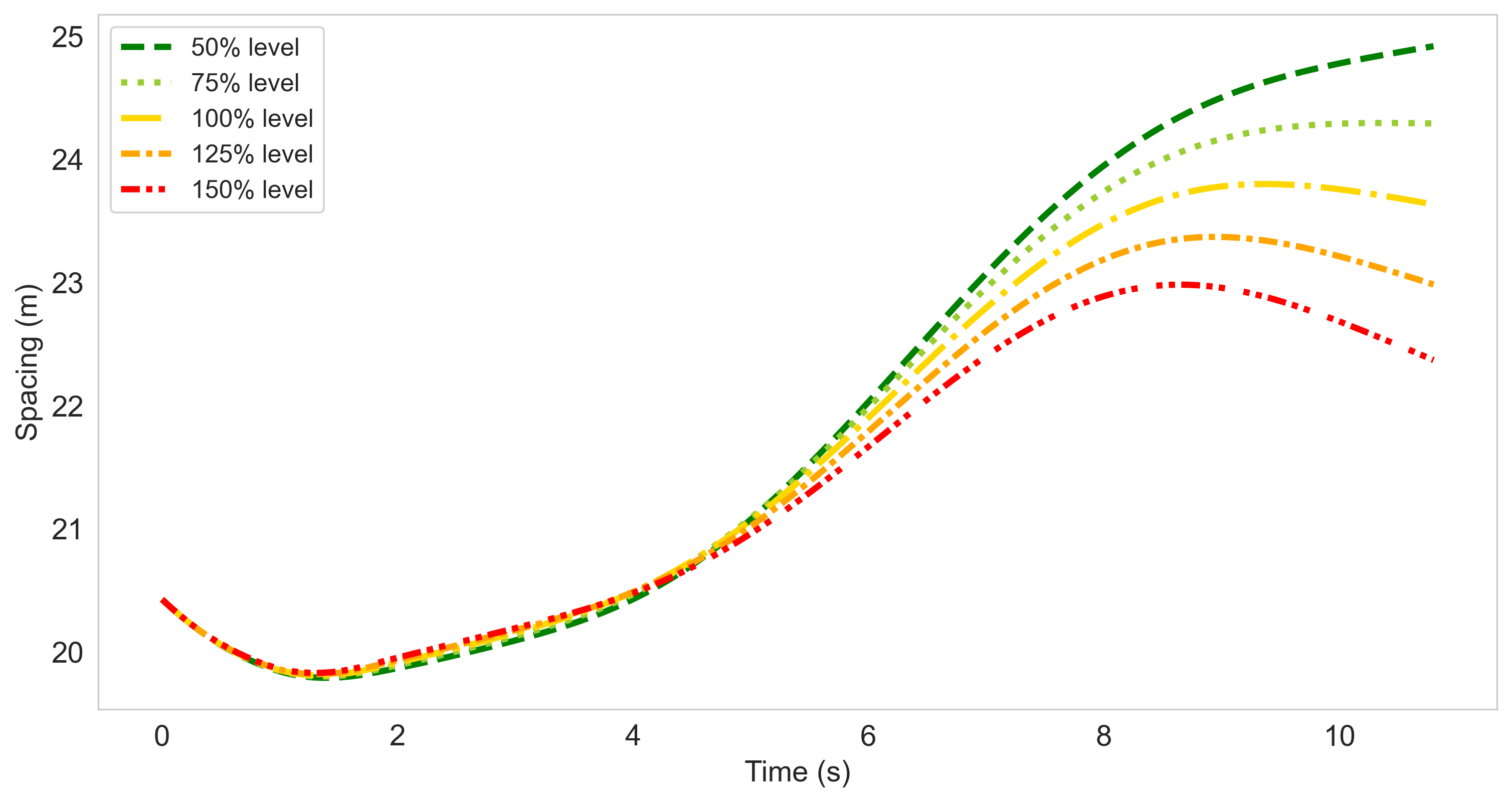}

\end{minipage}
\caption{Variations in car-following behavior with discourtesy levels: case(b).}
\label{fig:b}
\end{figure*}\begin{figure*}[htbp]
\centering
\begin{minipage}[t]{0.48\textwidth}
\centering
\includegraphics[width = \textwidth]{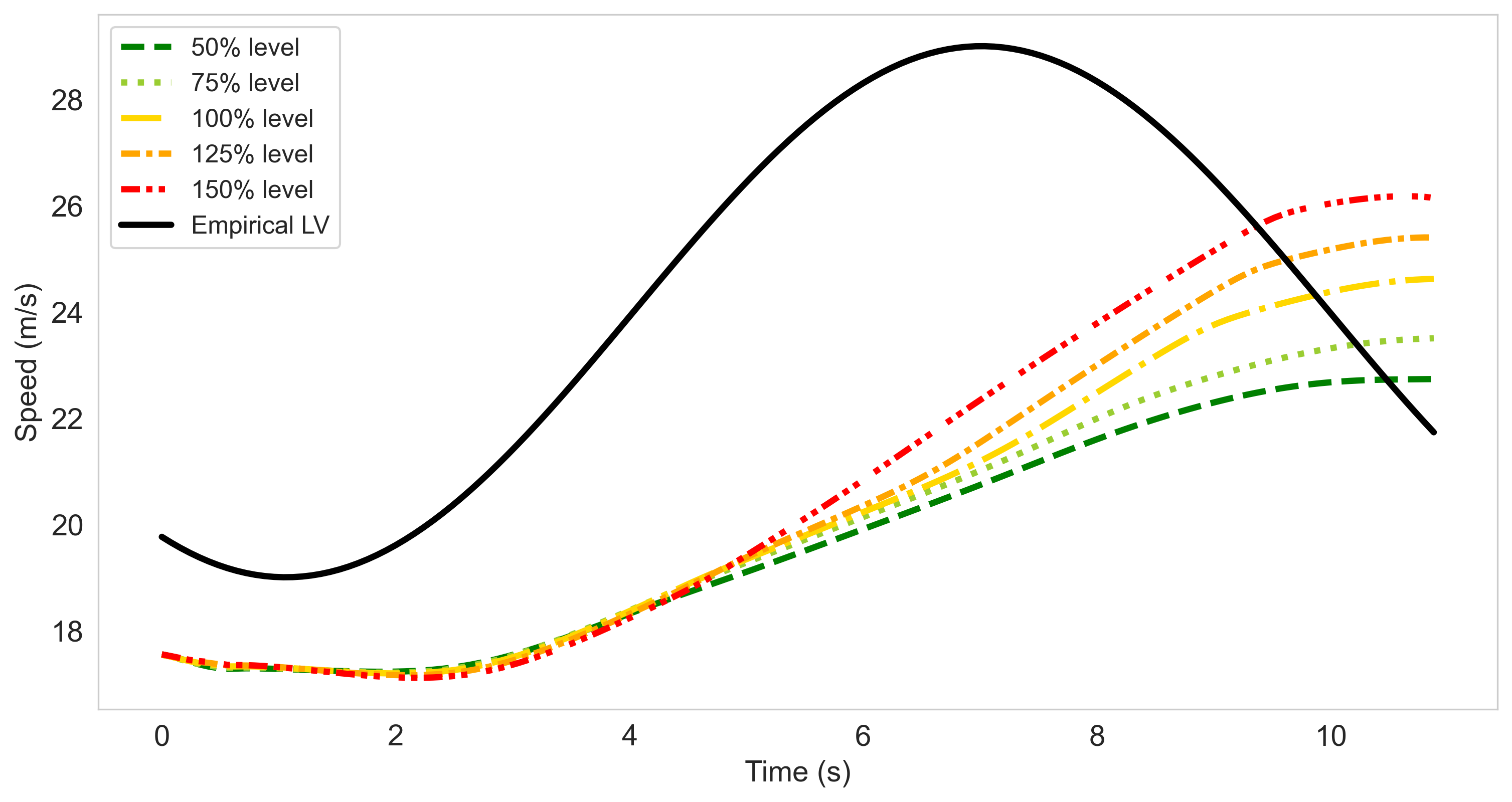}
\end{minipage}
\begin{minipage}[t]{0.48\textwidth}
\centering
\includegraphics[width = \textwidth]{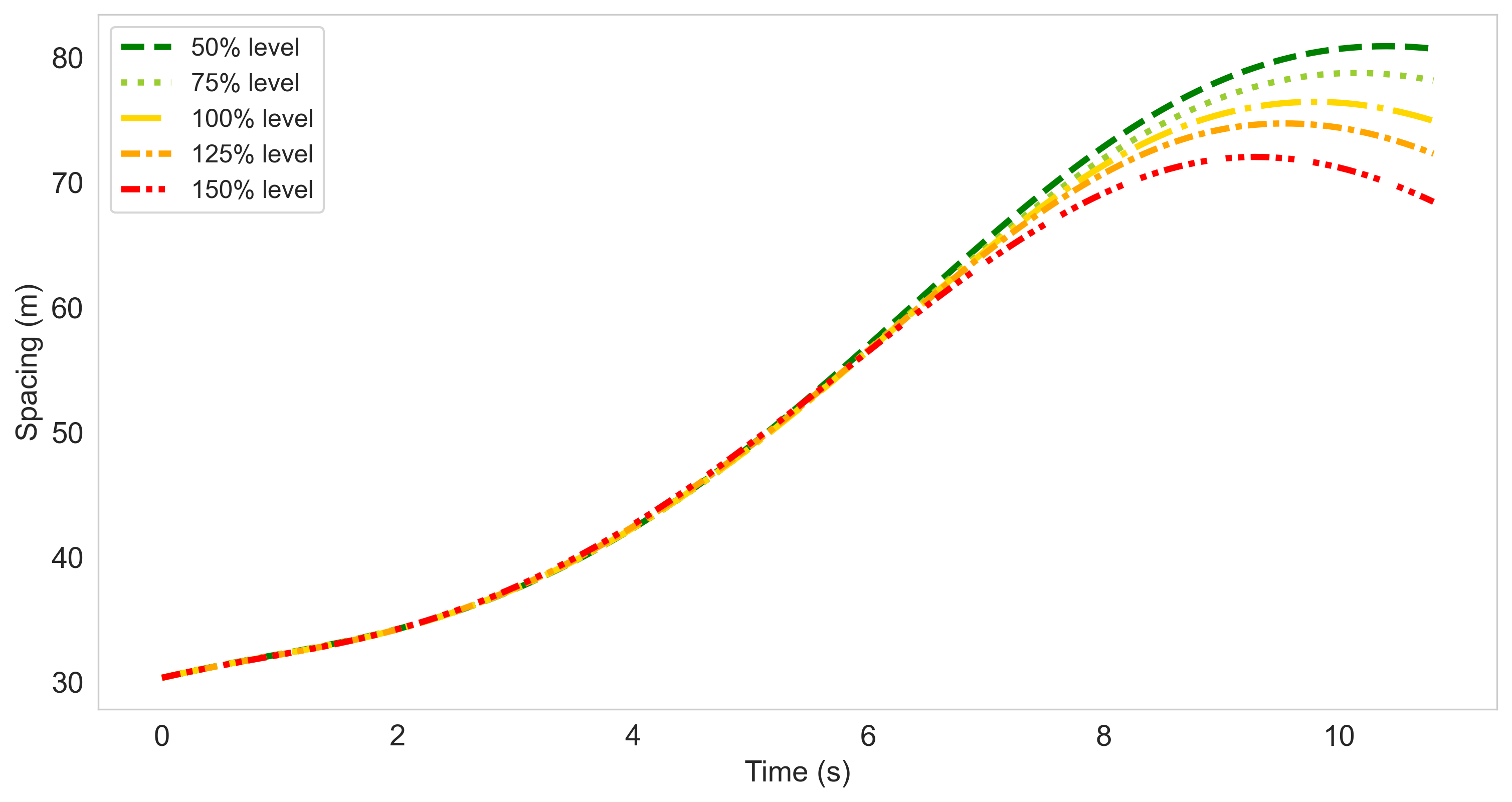}

\end{minipage}
\caption{Variations in car-following behavior with discourtesy levels: case(c).}
\label{fig:4}
\end{figure*}
    
\end{enumerate}

\subsection{Insights for adaptive cruise control development} 
Current ACC systems often employ fixed following distances, providing drivers with limited options, such as selecting long, medium, or short following distances. However, drivers have their own preferences, and maintaining a fixed distance may lead to challenges such as cut-ins when following too closely. Introducing a continuous and adjustable discourtesy level in ACC systems would enhance user experience and increase system penetration. Drivers could customize their driving preferences, leading to improved ACC adoption and overall driving satisfaction.

\section{Conclusion}
In this study, we introduced the Editable Behavior Generation (EBG) model, the first data-driven car-following model capable of adjusting driving discourtesy levels. Our model, leveraging diverse discourtesy calculation methods and integrating them into state-of-the-art data-driven car-following models, demonstrated competitive performance in terms of Spacing MSE, Speed MSE, and Courtesy MSE. The controllability of driving behavior was investigated by adjusting discourtesy levels, revealing insights into the model's adaptability and responsiveness. The ability of our model to align with drivers' social preferences opens new avenues for ACC design, potentially improving driver experience and system penetration rates.

\section{Future Work}
Future research endeavors may focus on several key aspects. Firstly, exploring additional courtesy calculation methods and their impact on driving behavior could further enhance the model's adaptability. Additionally, extending the evaluation to diverse datasets and driving scenarios would contribute to a more comprehensive understanding of the model's generalizability. Furthermore, conducting user studies to gather preferences and feedback on driving discourtesy levels from human drivers would facilitate the development of more socially aware models.

\section{Appendix}
Demonstration of car-following behavior under various discourtesy levels, please refer to the following video link: \url{https://www.youtube.com/watch?v=UWsF6lJE8jY}.

\bibliographystyle{IEEEtran}
\bibliography{ref}

\begin{IEEEbiography}[{\includegraphics[width=1in,height=1.25in,clip,keepaspectratio]{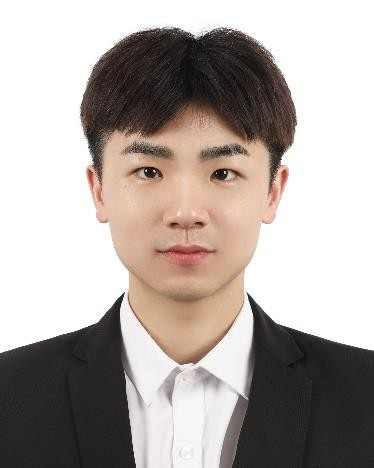}}]{Xianda Chen} received the M.S. degree from The University of Hong Kong, Hong Kong, China, in 2022. He is currently pursuing
a Ph.D. degree in intelligent transportation with The Hong Kong University of Science and Technology (Guangzhou), Guangzhou, China. His research interests include intelligent transportation, machine learning, and data analytics.
\end{IEEEbiography}

\begin{IEEEbiography}[{\includegraphics[width=1in,height=1.25in,clip,keepaspectratio]{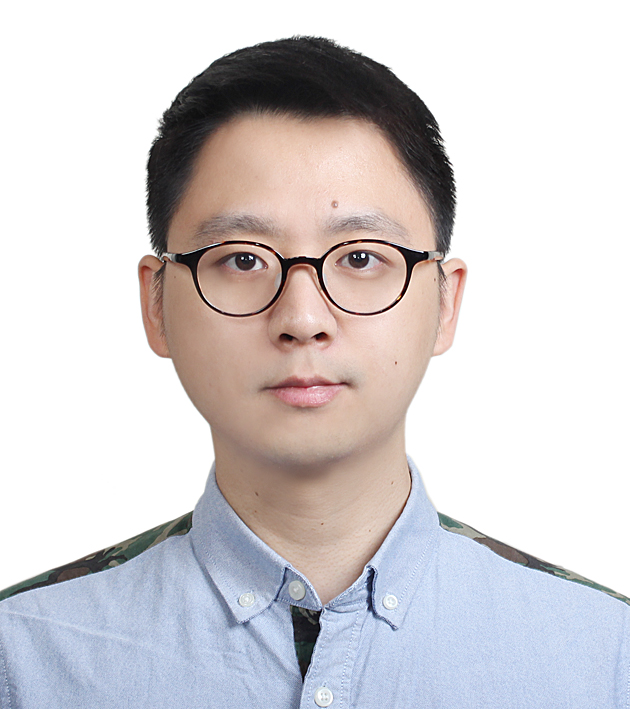}}]{Xu Han} received the M.Phil. degree in Internet of Things from the Hong Kong University of Science and Technology, Hong Kong, China, in 2015. He is currently pursuing a Ph.D. degree in Data Science and Analytics with the Hong Kong University of Science and Technology (Guangzhou), Guangzhou, China. His research interests include deep learning, reinforcement learning, decision intelligence and autonomous vehicles.
\end{IEEEbiography}

\begin{IEEEbiography}[{\includegraphics[width=1in,height=1.25in,clip,keepaspectratio]{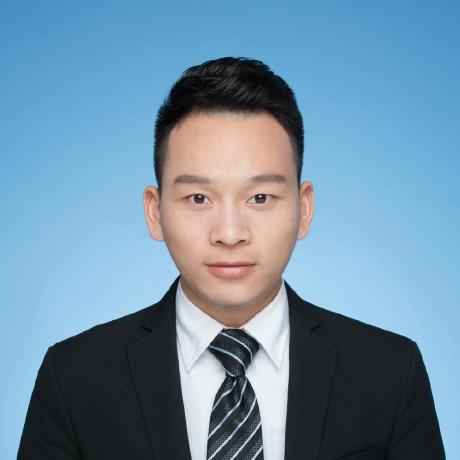}}]{Meixin Zhu} is a tenure-track Assistant Professor
in the Thrust of Intelligent Transportation
under the Systems Hub at the Hong Kong University
of Science and Technology (Guangzhou). He is also
an affiliated Assistant Professor in the Civil and
Environmental Engineering Department at the Hong
Kong University of Science and Technology. He
obtained a Ph.D. degree in intelligent transportation
at the University of Washington (UW) in 2022. He
received his BS and MS degrees in traffic engineering in 2015 and 2018, respectively, from Tongji
University. His research interests include Autonomous Driving Decision
Making and Planning, Driving Behavior Modeling, Traffic-Flow Modeling
and Simulation, Traffic Signal Control, and (Multi-Agent) Reinforcement
Learning. He is a recipient of the TRB Best Dissertation Award (AED50)
in 2023.
\end{IEEEbiography}

\begin{IEEEbiography}[{\includegraphics[width=1in,height=1.25in,clip,keepaspectratio]{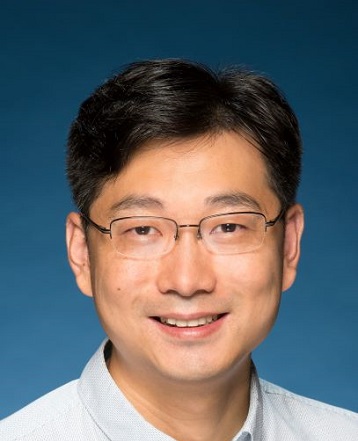}}]{Xiaowen Chu} is currently a Professor at the Data Science and Analytics Thrust, The Hong Kong University of Science and Technology (Guangzhou). Dr. Chu received his B.Eng. degree in Computer Science from Tsinghua University in 1999, and the Ph.D. degree in Computer Science from HKUST in 2003. He has been serving as the associate editor or guest editor of many international journals, including IEEE Transactions on Network Science and Engineering, IEEE Internet of Things Journal, IEEE Network, and IEEE Transactions on Industrial Informatics. He is a co-recipient of the Best Paper Award of IEEE INFOCOM 2021. His current research interests include GPU Computing, Distributed Machine Learning, and Wireless Networks.
\end{IEEEbiography}

\begin{IEEEbiography}
    [{\includegraphics[width=1in,height=1.25in,clip,keepaspectratio]{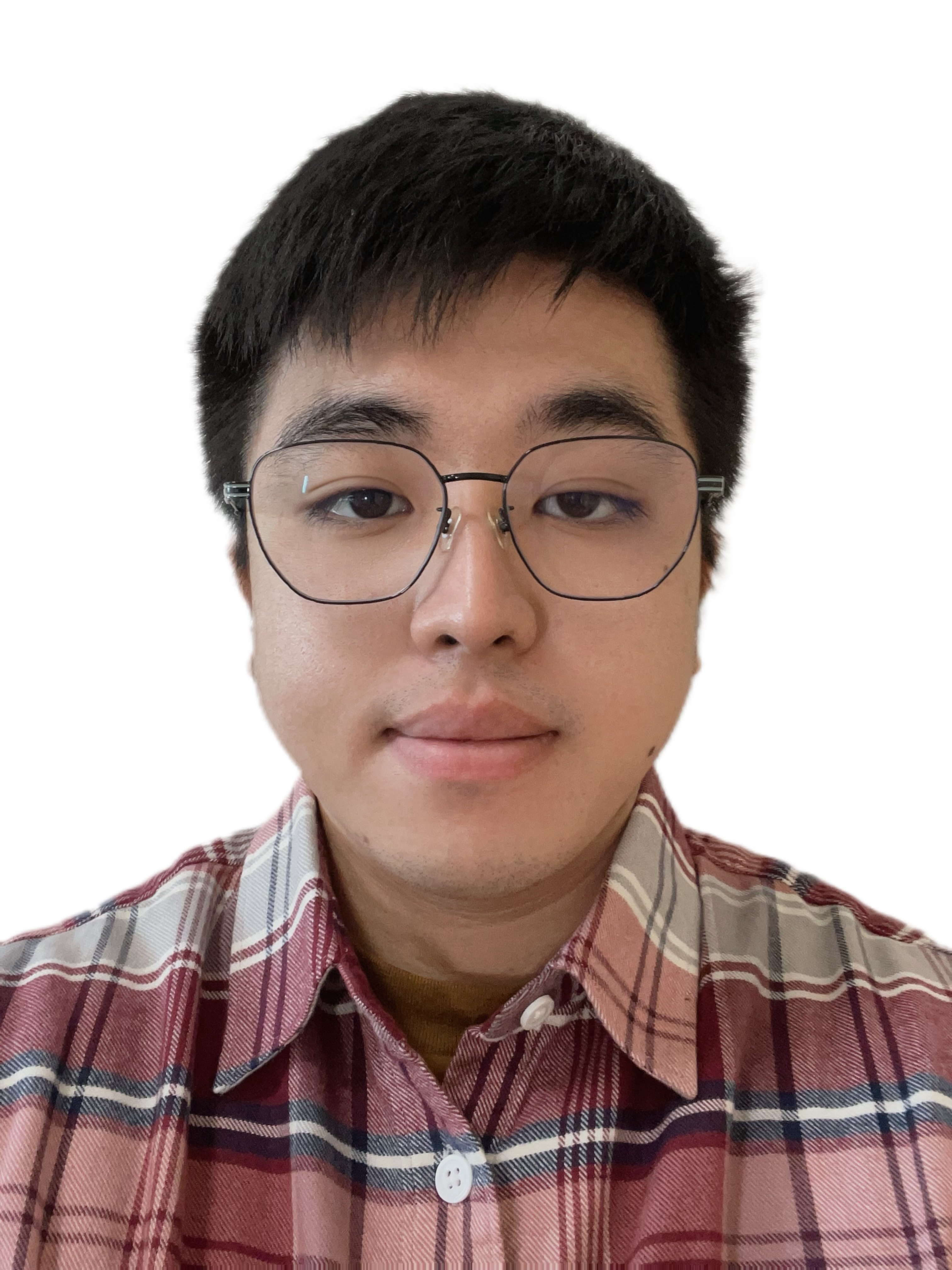}}]{PakHin Tiu}
    graduated with a Bachelor of Science degree in Applied Mathematics, Engineering, and Physics from the University of Wisconsin-Madison, United States, in 2020. He is currently pursuing a Master of Philosophy (MPhil) degree in Intelligent Transportation at the Hong Kong University of Science and Technology (Guangzhou), Guangzhou, China. His research interests encompass intelligent transportation and autonomous driving.
\end{IEEEbiography}

\begin{IEEEbiography}
[{\includegraphics[width=1in,height=1.25in,clip,keepaspectratio]{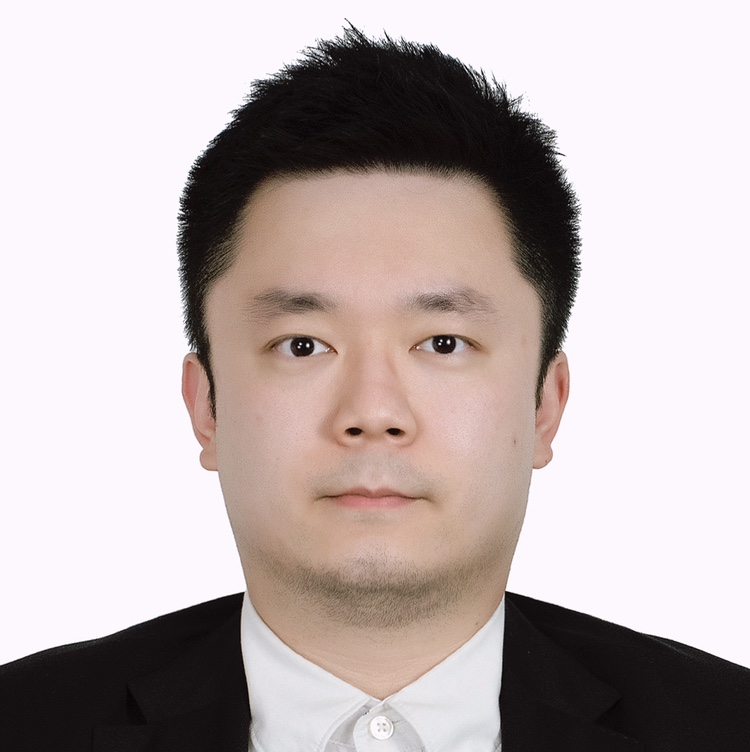}}]{Xinhu Zheng} received the Ph.D. degree in electrical and computer engineering from the University of
Minnesota, Minneapolis, in 2022. He is currently
an Assistant Professor with the Hong Kong University of Science and Technology (GZ). His current
research interests are data mining in power systems,
intelligent transportation system by exploiting different modality of data, leveraging optimization, and
machine learning techniques. 
\end{IEEEbiography}

\begin{IEEEbiography}[{\includegraphics[width=1in,height=1.25in,clip,keepaspectratio]{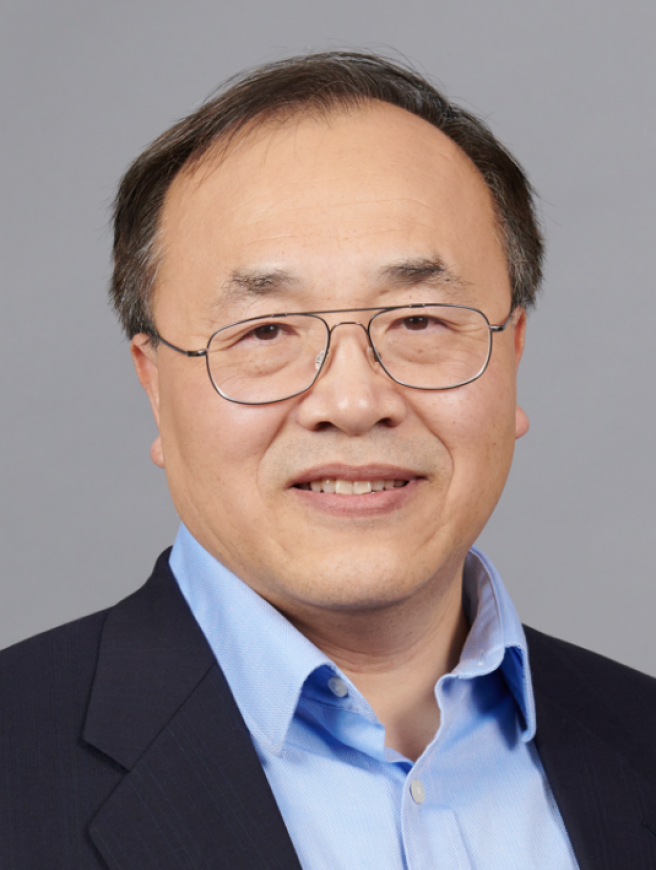}}]{Yinhai Wang}  (Fellow, IEEE) received the master’s
degree in computer science from the University of
Washington, (UW), Seattle, WA, USA, and the Ph.D.
degree in transportation engineering from The University of Tokyo, Tokyo, Japan, in 1998. He is currently a Professor of transportation engineering and
the Founding Director of the Smart Transportation
Applications and Research Laboratory (STAR Lab),
UW. His research interests include traffic sensing, urban mobility, e-science of transportation, and transportation safety. He is the Chair of the Artificial Intelligence and
Advanced Computing Committee of the Transportation Research Board. He is also a Member of the IEEE Smart Cities Technical Activities Committee. From 2010 to 2013, he was
an Elected Member of the Board of Governors for the IEEE ITS Society. He is also an Associate Editor for three journals, the Journal of Intelligent Transportation Systems, PLOS One, and Journal of Transportation Engineering.
\end{IEEEbiography}

\vfill

\end{document}